\documentclass{article}
\usepackage{iclr2027_conference,times}
\newif\ifpreprint
\iclrfinalcopy\preprinttrue 

\usepackage[utf8]{inputenc}
\usepackage[T1]{fontenc}
\usepackage{amsmath,amsfonts}
\usepackage{booktabs}
\usepackage{nicefrac}
\usepackage{microtype}
\usepackage{xcolor}
\usepackage{graphicx}
\usepackage{subcaption}
\usepackage{float}
\usepackage{hyperref}
\usepackage{url}
\usepackage{cleveref}
\hypersetup{pdftitle={How Do Language Models Choose Between Context and Memory?},
  pdfsubject={},pdfkeywords={}}
\ificlrfinal
  \hypersetup{pdfauthor={Benjamin Shih, John Winnicki, Arianna Cao}}
\else
  \hypersetup{pdfauthor={}}
\fi

\definecolor{ctteal}{HTML}{175E54}
\definecolor{ctcoral}{HTML}{E04F39}

\title{How Do Language Models Choose Between Context and Memory?}
\author{
Benjamin Shih\textsuperscript{1,2} \quad
John Winnicki\textsuperscript{1} \quad
Arianna Cao\textsuperscript{1}\\
\textsuperscript{1}Stanford University \qquad
\textsuperscript{2}Perpetual Labs\\
\texttt{\{benjamin.shih, winnicki, carianna\}@stanford.edu}
}

\begin{document}
\maketitle
\ifpreprint\lhead{Preprint.}\fi

\begin{abstract}
When contextual information conflicts with knowledge stored in model parameters, activation directions can be used to decode and steer which source the model follows.
However, successful steering does not establish that the unedited model uses those directions to choose between sources, or that they remain effective across tasks.
To test these possibilities, we vary the stated authority of contextual claims while holding their content fixed.
We first estimate authority directions from prompts in which context and parametric knowledge agree, then test their causal contribution when the two sources conflict.
Interchanging naturally occurring activation values along these directions between matched high- and low-authority prompts reproduces 30--68\% of the authority-induced shift in source choice across Qwen, Llama, and OLMo models, whereas matched controls reproduce almost none.
We next ask what transfers across tasks: the learned direction versus the activation values exchanged along it.
Using a direction learned on another task closed 9\% of the source-choice gap, compared with 57\% when learned on the task being evaluated.
Both interventions exchanged activation values from the evaluated task.
In a separate experiment, we kept its learned direction but exchanged values taken from another task, which closed 68\% of the gap.
These results show that authority-related activation values can causally influence source choice across tasks when inserted along directions learned for the task being evaluated.
\end{abstract}

\section{Introduction}
Language models can encounter contextual information that conflicts with knowledge stored in their weights, which we refer to as parametric memory \citep{spare}.
When these sources disagree, what determines which one the model follows?
We investigate how the stated authority of a contextual claim affects this choice, which internal representations contribute to that effect, and whether their influence transfers across tasks.

Prior work on context-memory arbitration has shown that language models contain 
low-dimensional representations that can control whether predictions follow contextual information or parametric knowledge. 
\citet{ccs} identify a one-dimensional context-sensitivity ``knob,''
\citet{spare} use sparse features to steer, and 
\citet{competition} trace how competing contextual and parametric mechanisms evolve through the network. 
More recent methods use activation steering or learned gates to control source choice \citep{contextfocus,shift}.
Complementary studies show that source authority systematically affects model responses \citep{whoendorsed}, with mechanistic analyses localizing an authority-related effect to a late layer \citep{authorityhier}.
Put together, these results demonstrate that source choice can be decoded and externally steered.

These findings leave open whether the unedited model uses the identified representations to choose between context and memory.
A representation may predict source choice, and interventions along it may change the answer, without establishing that its natural variation contributes to that choice.
\citet{actionability,languageaxis} examine this gap between representational accessibility and causal use,
and \citet{causalkey,causalityinvariance} illustrate the broader problem that these representations may not 
transfer across tasks or prompt formats.

We estimate authority directions from prompts in which context and memory agree on the answer.
We then exchange activation values between high- and low-authority conflict prompts to test whether variation along these directions contributes to source choice.

\paragraph{Our contributions are as follows:} 
\begin{enumerate}
    \item \textbf{Authority-related activations causally influence source choice.} Exchanging activation values between prompts that assign high or low authority to the same claim reproduces 30--68\% of the resulting shift between following context and memory.
    \item \textbf{Two forms of cross-task transfer.} We distinguish reusing a learned direction from transferring activation values. Directions have larger effects on the task used to estimate them, while values from another task can influence source choice when inserted along a direction learned for the receiving task.
    \item \textbf{Representing authority does not establish causal use.} Across tasks, these directions reliably encode authority, yet single-layer interventions along them have little effect on source choice.
\end{enumerate}
\section{Methodology}\label{sec:setup}
We construct prompts containing unambiguous answers to create scenarios where the model's
parametric memory supports a true fact while the context states a false fact, and format it to end with a question and answer prefix. 
To illustrate the mechanics of our method, we will use the following example color prompt:  
``\,\dots\ \texttt{What color is the grass? Answer: The grass
is}\,''. We compare the model's next-token scores (logits) over a fixed set of 12 colors. 
The complete color set and specifications are described in \Cref{app:color-task}, and we refer to this setup as the \emph{color task}. 

To test varying authority, we hold the false claim in the context fixed and change only how the prompt presents it. 
A high-authority prompt says, ``\texttt{The grass is definitely purple.}''; a low-authority
prompt says, ``\texttt{There is an unverified rumor that the grass is purple.}'' We also experiment 
over five authority cue families: authority wording, source credibility, cited evidence, consensus, and first-person confidence as an empirical null baseline (\Cref{app:authority-controls}).

We study four frozen, open-weight instruction-tuned models: Qwen2.5 (3B and 7B), Llama-3.1-8B, and OLMo-2-7B. 
The intervention layers are selected once in Qwen2.5-7B and the corresponding relative depth is used in the other models. 
No model is fine-tuned.

\begin{figure}[tb]
\centering
\includegraphics[width=0.80\linewidth]{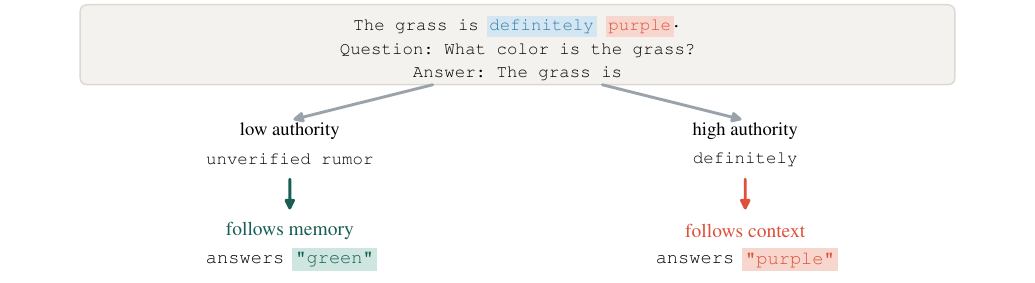}
\caption{The context asserts 
\textcolor{ctcoral}{purple}, which conflicts with the model memory supporting \textcolor{ctteal}{green} for the color task. The model response varies across high and low-authority framings.}
\label{fig:task}
\end{figure}

\subsection{Authority direction steering}
We first identify how the stated authority of a claim changes the model's activations while holding the correct answer fixed.
For the color task, we construct matched high- and low-authority prompts in which context and memory agree.
We use $\mathrm{H}$ and $\mathrm{L}$ to denote the high- and low-authority conditions, respectively.
At each layer $\ell$, we measure the residual-stream activation at the final prompt position and compute the mean activations $\mu_{\mathrm{H},\ell}$ and $\mu_{\mathrm{L},\ell}$.
Their difference defines the authority vector
\[
  d_\ell=\mu_{\mathrm{H},\ell}-\mu_{\mathrm{L},\ell},
  \qquad
  s_\ell=\frac{d_\ell}{\lVert d_\ell\rVert_2}.
\]

We then test whether changing activations along this vector alters source choice when context and memory conflict.
We add $\alpha d_\ell$ to low-authority prompts and subtract it from high-authority prompts, where $\alpha$ controls the intervention strength.
As a control, we apply the same intervention using a random vector matched to the norm of $d_\ell$.

To assess whether steering favors the answer supplied by context rather than a particular color, we measure logit changes across all twelve candidate answers.
We also test the stability of the estimated direction at layer 20 by repeating the estimation on ten randomly sampled halves of the factual subjects and measuring pairwise cosine similarity.

\subsection{Causal mediation of source selection}
Additive steering tests whether an authority vector can influence source choice.
We next ask whether naturally occurring variation along that direction contributes to the effect of authority.
For each matched pair of high- and low-authority conflict prompts, we exchange the activation component along $s_\ell$ while preserving the recipient's orthogonal component.

Let $h_\ell$ denote the residual-stream activation and $z_\ell=s_\ell^\top h_\ell$ its authority coordinate.
To give a low-authority prompt the coordinate from its high-authority counterpart, we apply
\[
  h'_{\mathrm{L},\ell}
  =h_{\mathrm{L},\ell}
  +\bigl(z_{\mathrm{H},\ell}-z_{\mathrm{L},\ell}\bigr)s_\ell.
\]
We also perform the reciprocal intervention, inserting the low-authority coordinate into the high-authority prompt.
In both cases, we take donor coordinates from unedited runs and repeat the replacement across eight layers, recomputing downstream activations after each edit.

We measure the resulting change in the fraction of prompts on which the model selects the contextual answer.
Let $p_{\mathrm{H}}$ and $p_{\mathrm{L}}$ denote these rates before intervention, and let $p_{\mathrm{L}\leftarrow\mathrm{H}}$ and $p_{\mathrm{H}\leftarrow\mathrm{L}}$ denote the rates after the two reciprocal interventions.
We report their combined effect relative to the original authority gap:
\[
  \mathrm{GapClosure}
  =\frac{(p_{\mathrm{L}\leftarrow\mathrm{H}}-p_{\mathrm{L}})
  +(p_{\mathrm{H}}-p_{\mathrm{H}\leftarrow\mathrm{L}})}
  {p_{\mathrm{H}}-p_{\mathrm{L}}}.
\]
The mediation index averages the two effects before normalization and therefore equals half the gap closure.
To assess whether the effect is specific to the authority direction, we repeat the same swaps using random directions and directions matched to the residual variation, keeping the prompt pairs and intervention sites fixed.

\subsection{Transferring authority values across tasks}
\label{sec:affine}
We next investigate whether the authority coordinates measured on one task can influence source choice in other tasks.
To do this, we introduce a new task, the \emph{material task}, that asks what objects are made of, with answers distinct from the color vocabulary.
As before, we estimate the authority directions from prompts in which the context and memory agree before replacing the coordinates along the material authority directions with values measured from color task prompts of the opposite authority.
At each layer, we fit a linear rescaling and offset between color and material coordinates using agreement prompts from both tasks. The material task supplies the intervention directions and calibration.

\begin{figure}[tb]
\centering
\begin{subfigure}[tb]{0.46\linewidth}\centering
\includegraphics[width=\linewidth]{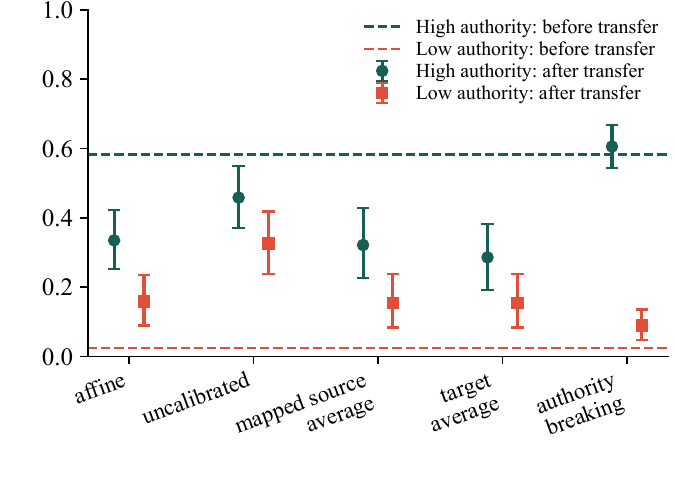}
\subcaption{Context following frequency after transfer.}
\end{subfigure}\hfill
\begin{subfigure}[tb]{0.46\linewidth}\centering
\includegraphics[width=\linewidth]{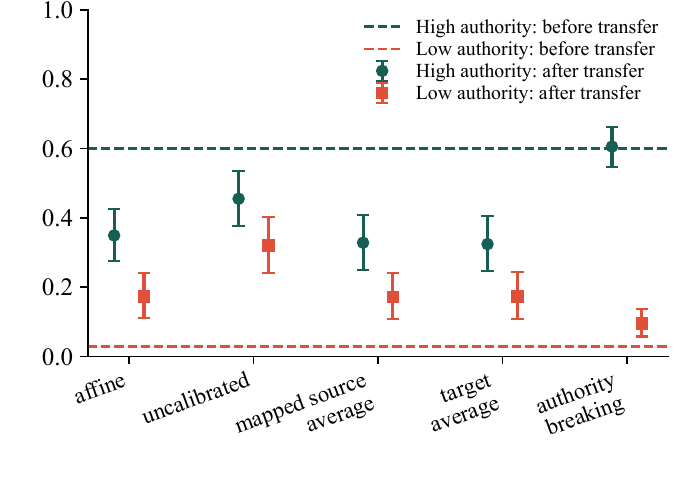}
\subcaption{Context answer probability after transfer.}
\end{subfigure}

\medskip
\begin{subfigure}[tb]{0.46\linewidth}\centering
\includegraphics[width=\linewidth]{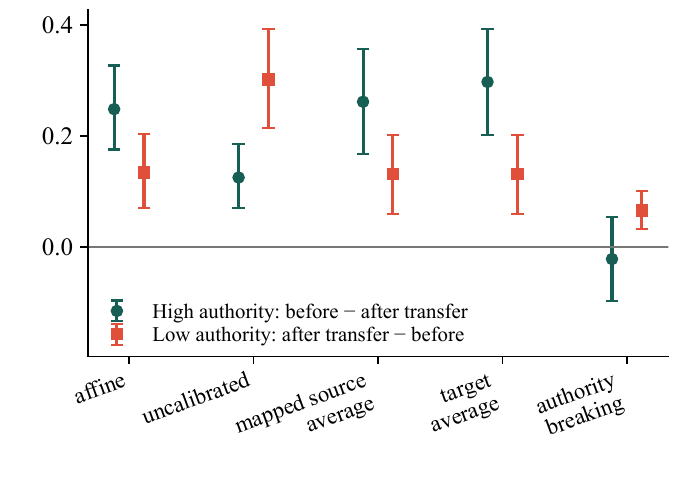}
\subcaption{Changes in context following frequency.}
\end{subfigure}\hfill
\begin{subfigure}[tb]{0.46\linewidth}\centering
\includegraphics[width=\linewidth]{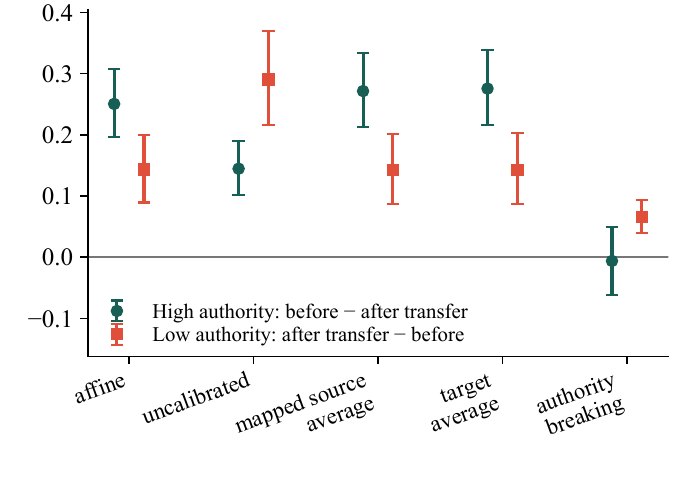}
\subcaption{Changes in context answer probability.}
\end{subfigure}
\caption{\textbf{Authority-value transfer from colors to materials.} Panels (a--b) show context-following frequency and answer probability; (c--d) show reductions for high-authority recipients and increases for low-authority recipients. Dashed lines mark unedited baselines. Probabilities are normalized over twelve material answers, and error bars show pointwise 95\% paired bootstrap intervals with maps and donors fixed.}
\label{fig:affine-transfer}
\end{figure}

There are three questions that this comparison poses. If the donor scalar carries authority information, the transfer should close the authority gap, and breaking the authority assignment should eliminate the effect. 
If affine calibration matters, the affine transfer should outperform the unaltered transfer. 
If the effect depends on information specific to individual donor prompts, then we expect their coordinate values to produce greater gap closure than the corresponding high or low authority averages at each layer. 
Only the first question received a clear answer. Affine donor transfer closed 0.684 of the source choice gap (95\% CI [0.546, 0.827]; \Cref{fig:affine-transfer}), whereas the authority-breaking control closed only 0.077 (paired difference 0.606 [0.468, 0.798]). The affine calibration did not help. Uncalibrated transfer closed 0.763, and their difference was -0.080 [-0.221, 0.042]. 
Nor did individual donors outperform the previously computed averages. Using the average color coordinate closed 70.2\% of the gap and using the average material coordinate closed 76.6\%, compared with 68.4\% for individual donors with affine calibration.                                                
So, individual donors did not show a clear advantage over averages.

The aggregate score hides two different effects. Compared with uncalibrated transfer, high authority recipients were less likely to follow context by 10.58 percentage points [8.52, 12.64].
This strengthened the intended effect of giving them a low authority signal. However this also made low authority recipients less likely to follow context by 14.75 points ([10.57, 19.20]), which weakens the intended effect of a high authority signal.
The two effects pull against each other, which is consistent with the lack of an aggregate benefit. 

These results support cross-task transfer of authority values when the receiving task supplies the intervention directions, without a clear advantage from calibration or individual donor values. Full details appear in \Cref{app:affine}.

\subsection{Additional coordinates generalize without authority specificity}
\label{sec:heldout-rank}
We next tested whether adding coordinates improves mediation on unseen color facts by testing subspaces of ranks 1 through 4 at each intervention layer. Here, we fit all bases on 36 subjects before opening 24 previously untouched test subjects in Qwen2.5-7B. 

At each of the eight layers, the first direction was the mean high minus low authority difference on training agreement prompts. We centered the training conflict differences, took their right singular vectors, then applied Gram-Schmidt against the agreement direction and previously accepted additions. The first three usable additions defined nested ranks one through four.  Each intervention copied the matched natural donor projection in both authority directions, recomputing the recipient trajectory after each site.

\begin{figure}[tb]
\centering
\begin{subfigure}[tb]{0.46\linewidth}\centering
\includegraphics[width=\linewidth]{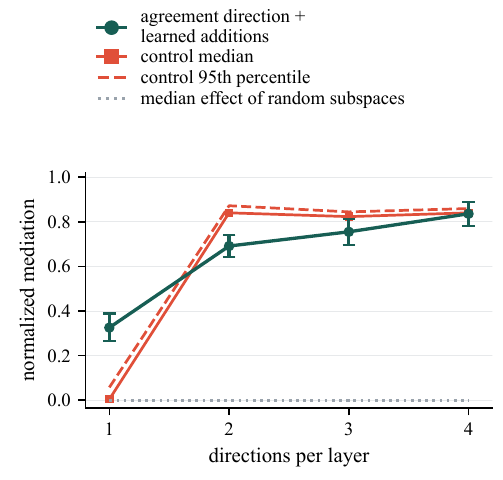}
\subcaption{Shuffled authority labels.}
\end{subfigure}\hfill
\begin{subfigure}[tb]{0.46\linewidth}\centering
\includegraphics[width=\linewidth]{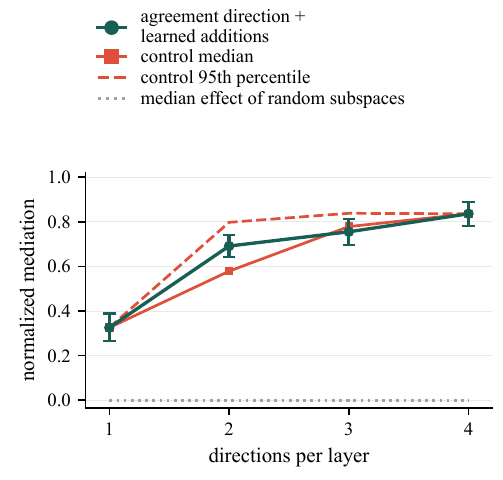}
\subcaption{Agreement direction + randomized additions.}
\end{subfigure}
\caption{\textbf{Additional coordinates improve mediation but fail authority-specificity controls.} (a) Mediation on held-out color facts as the number of intervention coordinates increases, compared with controls fitted using shuffled authority labels. (b) The same comparison with the agreement-derived authority direction held fixed and the additional directions randomized.}
\label{fig:heldout-rank}
\end{figure}

Let normalized mediation be the average of the two swap effects divided by the natural authority gap $G$: $M_k=(C_k^{\downarrow}+C_k^{\uparrow})/(2G)$. It is half the gap closure used for material transfer. On test subjects, $G=0.890$ and $M_k$ was 0.326, 0.691, 0.755, and 0.836 at ranks one through four (\Cref{fig:heldout-rank}). Rank two recovered 0.827 of the rank four effect (Fieller 95\% CI [0.754, 0.900]).

These effects generalized, but the added coordinates failed specificity controls. Each control family had 99 fixed draws. Shuffled supervision reassigned individual activation rows to balanced pseudo authority classes within color blocks, then rebuilt the basis using new pairings across subjects. We also added 3 rotated additional directions to the original agreement direction within the training rank four span. Shuffled ranks two and four met or exceeded the target in 99/99 and 66/99 draws, respectively. The target rank two effect also fell below the matched prefix 95th percentile. Thus, the result supports causal efficacy of additional conflict coordinates on unseen subjects, but not their specificity to correctly supervised authority (\Cref{app:heldout-rank}).

\section{Authority directions steer source choice across answers and cues}
\label{sec:authority-steering-cues}

We estimate authority directions from agreement prompts, where context and
memory support the same answer, and evaluate steering on conflict prompts.
The color task balances answer identities so that authority is not
systematically associated with a particular color.

\subsection{Steering across answer identities}

\begin{figure}[tb]
\centering
\begin{subfigure}[tb]{0.46\linewidth}
\centering
\includegraphics[width=\linewidth]{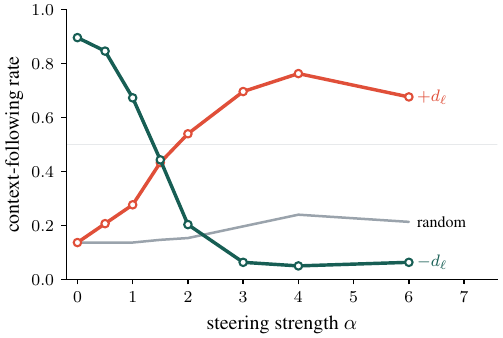}
\subcaption{Additive steering effect.}
\label{fig:dose}
\end{subfigure}
\hfill
\begin{subfigure}[tb]{0.46\linewidth}
\centering
\includegraphics[width=\linewidth]{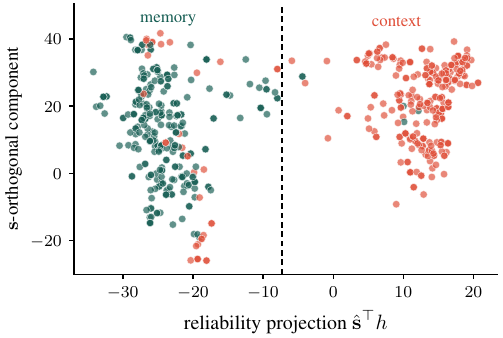}
\subcaption{Readout geometry.}
\label{fig:predict}
\end{subfigure}
\caption{\textbf{Authority directions predict and steer source choice.} In Qwen2.5-7B, (a) adding or subtracting the authority vector shifts context following more than a norm-matched random vector, and (b) the authority projection separates context-following from memory-following predictions.}
\label{fig:authority-steering-readout}\label{fig:steer}
\end{figure}

At layer index 21 in Qwen2.5-7B, adding the authority direction to low-authority
conflict prompts at $\alpha=4$ increases context following from 0.14 to 0.76. Subtracting
the direction from high-authority prompts at $\alpha=3$ reduces context following from
0.90 to 0.06. \Cref{fig:authority-steering-readout}a shows the
responses over the tested steering strengths, together with the smaller
effect of a norm-matched random direction. These interventions can
increase or decrease the rate at which the model follows the contextual
claim.

To check whether the effect tracks the asserted answer, we compare logit
changes across all twelve color candidates. The context-backed answer
receives the largest logit increase on 76\% of examples. Re-estimating
the direction at layer index 20 on ten random half-subsets of factual
subjects (which can overlap across estimates) also gives closely aligned
directions, with mean pairwise cosine similarity 0.96 and minimum 0.93.
These checks provide evidence against a steering effect driven only by
a fixed preference for one color. They show that the direction can favor
the answer supplied by the context across different answer identities.

The authority projection separates predictions favoring context over memory
more clearly than the displayed orthogonal component
(\Cref{fig:authority-steering-readout}b). Together, readout and additive
steering establish that the representation is informative and can influence
the output. Whether its natural variation contributes to unedited source
choice requires the coordinate-interchange test below.

\subsection{Transfer across authority cues}\label{app:authority-controls}

We next test transfer among authority wording, source credibility, cited
evidence, consensus, and first-person confidence. For each family, we
estimate a direction and test steering on the other families, keeping
the color task and answer vocabulary fixed.

\begin{figure}[tb]
\centering
\includegraphics[width=0.62\linewidth]{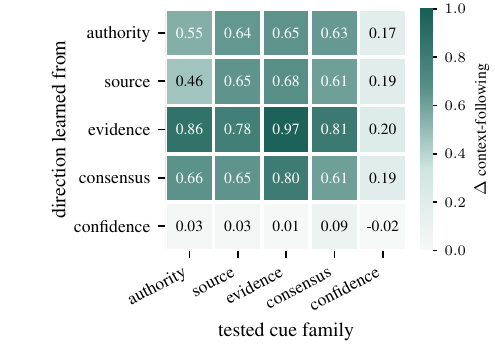}
\caption{\textbf{Authority steering transfers across cue families.} Rows identify the cue family used to estimate each direction; columns identify the family evaluated. Cells show the increase in context following on low-authority color prompts in Qwen2.5-7B at layer 21, with steering strength $\alpha=3$.}
\label{fig:authority-cue-transfer}\label{fig:cue}
\end{figure}

Directions estimated from wording, source credibility, evidence, and
consensus cues increase context following across the responsive cue
families. This pattern occurs in both Qwen and Llama, with effects
substantially larger than those of a random direction. The cross-cue
comparison in \Cref{fig:authority-cue-transfer} shows that the
effect extends beyond the cue family used for estimation. In particular,
the results are not confined to the ``definitely'' versus ``unverified
rumor'' contrast used in the running example.

In Qwen, first-person confidence behaves differently. Changing ``I might be
wrong'' to ``I am certain'' changes natural behavior by only about 0.01,
and the direction estimated from this contrast has little steering
effect across the tested families. This observation concerns the
confidence contrast and its learned direction: directions estimated
from the other families can still shift predictions on confidence
prompts. Cross-cue transfer therefore holds for several authority
manipulations within the color task, while providing a separate test
from direct reuse of a direction across factual tasks.

\section{Natural authority coordinates causally mediate source choice}\label{sec:local}
\begin{figure}[tb]
\centering
\begin{subfigure}[tb]{0.46\linewidth}\centering
\includegraphics[width=\linewidth]{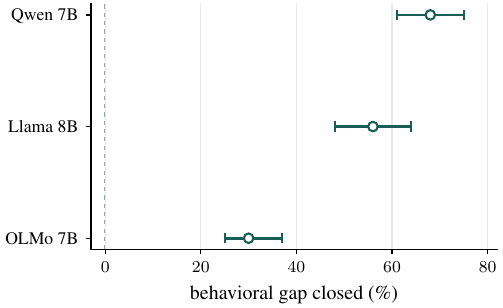}
\subcaption{Color task across models.}\label{fig:boundarylocal}
\end{subfigure}\hfill
\begin{subfigure}[tb]{0.46\linewidth}\centering
\includegraphics[width=\linewidth]{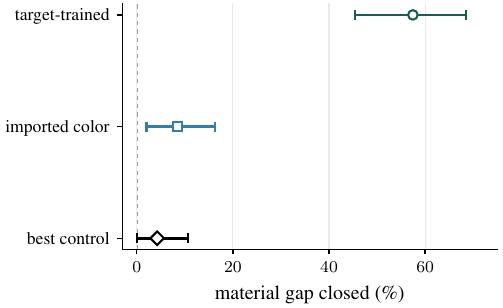}
\subcaption{Material task across direction sources.}\label{fig:boundarytransfer}
\end{subfigure}
\caption{\textbf{Interchanging authority coordinates changes source choice within and across tasks.} (a) Fraction of the color-task authority gap closed across models. (b) Fraction of the material-task gap closed using material-trained directions, color-trained directions, or the strongest matched control. All interventions in (b) exchange activation values from material prompts.}
\label{fig:boundary}
\end{figure}

\begin{figure}[tb]
\centering
\begin{subfigure}[tb]{0.46\linewidth}\centering
\includegraphics[width=\linewidth]{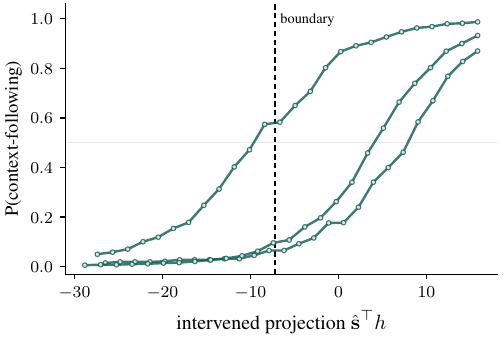}
\subcaption{Interventions per prompt.}\label{fig:medbars}
\end{subfigure}\hfill
\begin{subfigure}[tb]{0.46\linewidth}\centering
\includegraphics[width=\linewidth]{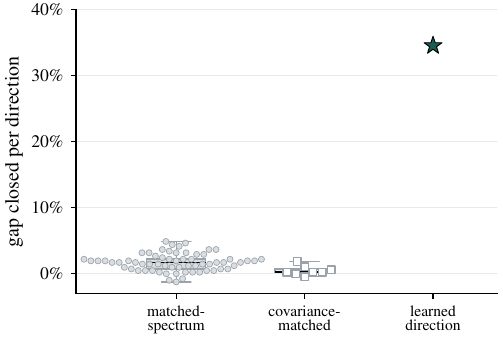}
\subcaption{Authority intervention versus matched controls.}\label{fig:swarm}
\end{subfigure}
\caption{\textbf{Causal effects of authority-coordinate interventions in Qwen2.5-7B.} (a) Probability of following context as the authority coordinate is varied on color prompts; the dashed line marks the decision threshold between context and memory. (b) Mediation from exchanging natural authority coordinates between matched prompts, compared with matched control directions.}
\label{fig:specificity}
\end{figure}

In the color task, interchanging the authority coordinates changes source choice. It closes 68\% of the natural 
high--low authority gap in Qwen, 56\% in Llama, and 30\% in OLMo; random interchange is near 
zero (\Cref{fig:boundary}). This provides evidence that the intervention mediates 
source choice within a task under repeated interchange.
Since a swap at 
a single site has a smaller effect, we sweep this over multiple layers to test a distributed late-layer direction set. As such, the effect is strongest in Qwen where the layer-specific direction set produces greater gap closure than the controls under the same intervention (\Cref{fig:specificity}). The complete control 
descriptions and model-specific layer intervals are in \Cref{app:models}.

\section{Limited direct cross-task direction reuse}\label{sec:transfer}
We construct another task, containing facts about what objects are made of, such that the next-token answers are disjoint from the color vocabulary. Keeping the prompt structure, intervention layers, scoring, and repeated interchange unchanged, memory accuracy is 0.988, and the natural authority gap is 0.560 (95\% CI [0.476, 0.643]). Importing the color direction set while interchanging natural material-donor coordinates along that basis closes 0.085 ([0.020, 0.163]), versus 0.574 ([0.455, 0.684]) for the native material direction: 14.8\% of the native effect. The strongest of the eight covariance-matched controls closes 0.043, and the imported color trajectory outperforms the strongest control by roughly 0.043. Thus direct reuse of the color basis is attenuated despite a strong native authority effect; this test does not transfer actual color-donor activations. Full details and controls are in \Cref{app:material-task}.

Directions learned on the evaluated task have larger effects in both
transfer directions. On color questions, material-derived directions close
21.4\% of the authority gap, compared with 73.6\% for color-derived
directions. Both paired differences have 95\% bootstrap confidence intervals
above zero (\Cref{appxtransfer:task-reuse}).

\section{Authority representation is not causal use}\label{sec:scope}
We next test whether an authority direction that is decodable across tasks also retains causal influence on source choice. To do this, we transfer the authority direction between state and country facts, for which we verify that the model knows the underlying fact without context and that the authority gap exists (\Cref{app:state-country}). We evaluate both directions by fixing an authority direction from one domain and evaluating it on the other across all 28 decoder layers.

For each example, we compare the ordinary reference prompt ($R$) with a control ($EA0$) that adds statements identifying each entity with itself and instructing the model to output each candidate answer unchanged. These additions preserve the task's factual content and answer mapping, allowing us to test sensitivity to redundant prompt wording. Under \(R\), authority changes behavior by 0.454 for countries and 0.492 for states; under \(EA0\), these gaps fall to 0.120 and 0.065. The authority gap remains decodable: state/country AUROC is 0.995--1.000 and country/state AUROC is 1.000; the imported authority     
 direction has negligible effect: the largest effects across all layers are 0.014 in the state/country direction and 0.038 in the reverse direction (\Cref{tab:transfernull}). Thus, authority can remain highly represented across state/country domains, while the intervention has negligible effect.

   \begin{table}[tb]
\centering
\caption{Cross-task authority decoding and single-layer intervention effects across all 28 decoder layers. Columns show the task used to estimate the direction followed by the task evaluated. $R$ denotes the reference prompt; $EA0$ adds redundant entity and answer mappings. Brackets give 95\% confidence intervals; parentheses identify the layer of the maximum effect.}
\label{tab:transfernull}
\small
\begin{tabular}{lcc}
\toprule
Measure & States $\to$ countries & Countries $\to$ states\\
\midrule
Authority gap ($R$) & 0.454 & 0.492\\
Authority gap ($EA0$) & 0.120 & 0.065\\
Authority AUROC & 0.995--1.000 & 1.000--1.000\\
Encoding retained & 0.921 [0.908, 0.934] & 0.900 [0.890, 0.911]\\
Maximum effect ($R$) & 0.014 [0.009, 0.019] (L20) & 0.038 [0.029, 0.047] (L17)\\
Maximum effect ($EA0$) & 0.011 [0.006, 0.016] (L17) & 0.026 [0.020, 0.033] (L18)\\
\bottomrule
\end{tabular}
\end{table}

As a robustness check, we incorporate an additional component in the activation to account for the change between $R$ and $EA0$ in the prompt, which does not restore a meaningful imported causal effect, with essentially null estimates -0.038 and -0.049. We also incorporate generic padding, which adds irrelevant additional text; this produces similar suppression, indicating that the null effect is not specific to the $EA0$ wording. 

These tests on state and country facts reveal that high cross-domain authority decoding does not necessarily translate to significant causal behavior changes. 

\section{Discussion}

\paragraph{Related work}
Prior work studies how models resolve knowledge conflicts and respond to source cues \citep{juice,tokenauthority,truthoverridden}.
Our experiments ask whether authority directions estimated without conflict contribute to source choice and remain effective across tasks.
This distinction matters because decodable information need not produce faithful explanations \citep{decodablefaithful}, and observable latent-state patterns need not causally affect behavior \citep{latentpatterns}.
Causal registers provide a complementary example in which later computation uses an edited intermediate state \citep{causalregisters}, while conditional-routing studies reveal limits to cross-task reuse \citep{testthenroute}.

Activation-source analyses and mechanistic evaluation frameworks further motivate matched controls and claims scoped to the tested intervention \citep{steeringsources,unit_tests,cif,mib,patchkernels}.
We apply these principles to source choice by separating additive steering, interchange of natural activation values, and cross-task reuse of learned directions.

\paragraph{Limitations and future work}
Our findings concern a limited set of tasks, prompts, and models.
Weak transfer may reflect differences in prompt sampling, activation geometry, or scale; it does not rule out a shared nonlinear representation with task-specific linear directions.
The state--country experiments test single-layer interventions, leaving repeated and higher-dimensional interchange in that setting unresolved.
Broader task coverage and comparisons of aligned task-specific subspaces could clarify when the observed limits reflect the intervention rather than the underlying computation.

\section{Conclusion}
\label{sec:conclusion}

Authority directions estimated from prompts where context and memory agree can steer source choice across answers and several cue families.
Interchanging natural activation values along these directions reproduces 30--68\% of the authority-induced shift across Qwen, Llama, and OLMo, supporting a causal contribution to source choice under the tested intervention.
This influence is less reusable across tasks: directions learned on the evaluated task have larger effects in both directions of the color--material comparison, while imported single-layer interventions have little effect on state--country source choice despite strong authority readout.
Repeated and higher-dimensional interventions remain open questions for the state--country setting.

Activation values from the color task can nevertheless influence material source choice when inserted along a material-trained direction.
This distinguishes transferring activation values from reusing a direction across tasks.
Neither affine calibration nor individual donors show a clear overall advantage over uncalibrated transfer or authority averages, and additional coordinates improve mediation on unseen color subjects without passing the tested authority-specificity controls.
Together, these findings establish a causal contribution of authority-related activations to source choice and show why cross-task transfer must be tested separately for the direction and the values exchanged along it.

\ificlrfinal
\subsubsection*{Acknowledgments}
Benjamin Shih acknowledges support from Perpetual Labs, including computational resources used in this work.
\fi

\section*{AI use statement}
We used generative AI tools to help implement and debug experiment and analysis code, edit manuscript prose, and prepare figures and tables. AI suggestions informed the work but did not constitute empirical evidence; reported measurements come from the model experiments described in this paper. The authors reviewed the AI-assisted work and take responsibility for the final text, methods, results, and claims.

\bibliographystyle{iclr2027_conference}
\bibliography{references}

\clearpage
\appendix
\crefalias{section}{appendix}
\crefalias{subsection}{appendix}
\crefalias{subsubsection}{appendix}

\section{Models}
\label{app:models}
 We use frozen Qwen2.5-7B-Instruct, Qwen2.5-3B-Instruct, Llama-3.1-8B-Instruct, and OLMo-2-7B-Instruct; no model is fine-tuned. Qwen2.5-3B is used only for readout and steering. 

\begin{table}[tb]
\centering
\caption{Models used and positions of the answer layer.}
\label{tab:models}
\small
\begin{tabular}{lccc}
\toprule
Model & Depth & Readout layer & Relative depth\\
\midrule
Qwen2.5-7B-Instruct & 28 & 20 & 0.71\\
Qwen2.5-3B-Instruct & 36 & 27 & 0.75\\
Llama-3.1-8B-Instruct & 32 & 23 & 0.72\\
OLMo-2-7B-Instruct & 32 & 24 & 0.75\\
\bottomrule
\end{tabular}
\end{table}

\subsection{Layer breakdown}
The color and material tasks swap activation components at layers $10,12,\ldots,24$ in Qwen2.5-7B and Llama-3.1-8B and relative-depth-matched layers $12,14,\ldots,26$ in OLMo-2-7B. Qwen2.5-3B is excluded from the multi-layer analysis, which makes sequential edits along these layers in a single run to see whether repeatedly exchanging this direction controls behavior. The state/country task sweeps all 28 decoder layers, performing single-layer interchange to identify whether a layer can control behavior alone. Covariance-matched controls are calculated by taking the difference between two real residual activations, removing the authority direction by projecting out the authority component and normalizing:
\[
\mathbf{c}_\ell
=
\frac{
P_\ell^\perp
\left(
h_{a,\ell}-h_{b,\ell}
\right)
}{
\left\|
P_\ell^\perp
\left(
h_{a,\ell}-h_{b,\ell}
\right)
\right\|_2
},
\qquad
P_\ell^\perp=I-\hat{\mathbf{s}}_\ell\hat{\mathbf{s}}_\ell^\top.
\]

Matched-spectrum controls preserve empirical singular-value structure and norm; they are formed by computing the SVD of a matrix of residual activations $H_\ell = U_\ell \Sigma_\ell V_\ell^\top$, scaling the coefficients by their empirical singular values
\[
\tilde{\mathbf r}_\ell^{(k)}
=
V_\ell \Sigma_\ell \mathbf g^{(k)},
\qquad
\mathbf g^{(k)} \sim \mathcal N(\mathbf 0,I),
\]
 and projecting out the authority direction and normalizing: 
\[
\mathbf r_\ell
=
\frac{
P_\ell^\perp \tilde{\mathbf r}_\ell^{(k)}
}{
\left\|
P_\ell^\perp \tilde{\mathbf r}_\ell^{(k)}
\right\|_2
},
\qquad
P_\ell^\perp
=
I-\hat{\mathbf s}_\ell\hat{\mathbf s}_\ell^\top.
\]
Every control uses the same layers, donors, recipients, and sign convention.

\section{Tasks}
\subsection{The color task: authority steering}
\label{app:color-task}
For the color task, each prompt outputs a single-token as the answer chosen from a fixed set of twelve colors. The prompts are formulated in a way that the colors are balanced so no authority is systematically associated with any color. The authority direction $s_\ell$ is estimated using the mean residual activations of 200 prompts for each authority and evaluated on 400 authority-conflicting pairs. We bootstrap the paired prompts 5000 times, resampling the conflict pairs.

\begin{table}[tb]
\centering
\caption{Interchange results on the color task.}
\label{tab:color-results}
\small
\begin{tabular}{lcccc}
\toprule
& Qwen & Llama & OLMo & Random swap\\
\midrule
Gap closure (\% natural gap) & 68 & 56 & 30 & $\approx0$\\
Bootstrap 95\% CI & [61, 75] & [48, 64] & [25, 37] & ---\\
Margin over best covariance control ($\times$) & 18 & 2.9 & 2.2 & ---\\
\bottomrule
\end{tabular}
\end{table}

\subsubsection{Color results}


The additive-steering, answer-identity, and cue-family results are
reported in \Cref{sec:authority-steering-cues} and
\Cref{fig:authority-steering-readout,fig:authority-cue-transfer}.

In the separate answer-logit control run, adding the layer-21 direction
at $\alpha=3$ to 300 low-authority conflict prompts changes the context-backed
and memory-backed answer logits by +5.96 and -8.17 on average.

In Qwen, the mediation index is 0.35, versus covariance-control mean 0.017 and maximum 0.049; it is 7.0\texttimes{} the best matched-spectrum control ($p=0.015$). Llama and OLMo margins over their strongest covariance controls are 2.9\texttimes{} and 2.2\texttimes{}, while matched-spectrum separation is 1.2\texttimes{} in Llama and marginal in OLMo. Readout and edit effects rise after layer 16 and peak around layers 18--20; MLP writes are largest there but are not necessary.

\begin{figure}[tb]
\centering
\begin{subfigure}[tb]{0.46\linewidth}
\centering
\includegraphics[width=\linewidth]{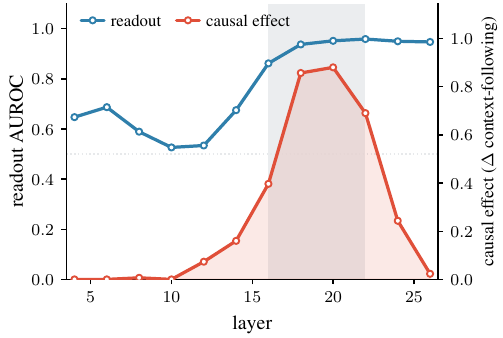}
\subcaption{Readout and edit effect by layer.}
\end{subfigure}
\hfill
\begin{subfigure}[tb]{0.46\linewidth}
\centering
\includegraphics[width=\linewidth]{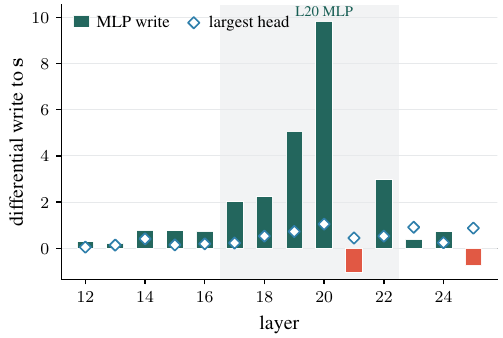}
\subcaption{Component writes.}
\label{fig:lmwrite}
\end{subfigure}
\caption{Late-layer behavior in Qwen2.5-7B on the color task.}
\label{fig:layers}
\end{figure}

We also performed dimension reduction on the authority subspace to test whether it occupies more than one dimension. Rank two recovers 0.827 of rank four (Fieller 95\% CI [0.754, 0.900]) but fails after shuffling labels and data.

\subsubsection{Held out rank construction}
\label{app:heldout-rank}
For the experiment in \Cref{sec:heldout-rank} 
, we used two controls. First, we shuffled authority labels, such that each row was assigned pseudo authority classes, grouping the agreement prompts and conflict prompts separately by their correct color and asserted color, respectively. Each group had a balanced number of high and low authority rows. We used four agreement directions to test whether our chosen second direction was more effective than the other directions in the same space. 

Averaging each subject's results across the false color claims, we calculated 95\% intervals using 20,000 bootstrap samples. We compared the average effect of the two swap directions against the 95\% of the control distribution and found that the rank 2 effect was lower than the randomly mixed direction by -0.095 [-0.131, -0.061] and the rank 4 effect was lower than the shuffled labels by -0.021 [-0.045, 0.000]. This indicates that the additional directions failed our test of authority specificity despite affecting behavior on new subjects. 

\subsection{The material task: limited causal transferability}
\label{app:material-task}
This task contains 132 objects (11 objects per material); descriptions omit the true material and mappings were derived from a fixed Wikipedia revision. The materials were assigned to a 4/7 train-test split (48/84 for objects). We performed 4000 bootstrap replications. As a control, we use 8 covariance-matched directions and copy the natural activation component to itself as a sanity check.

\begin{table}[tb]
\centering
\caption{Ratio of imported color closure over natural material closure under repeated interchange on the material task.}
\label{tab:material-results}
\small
\begin{tabular}{lc}
\toprule
Quantity & Estimate\\
\midrule
Memory accuracy & 0.988\\
Natural authority gap & 0.560 [0.476, 0.643]\\
Natural material closure & 0.574 [0.455, 0.684]\\
Imported-color closure & 0.085 [0.020, 0.163]\\
Imported/natural closure ratio & 0.148\\
Strongest covariance-control closure & 0.043\\
Imported minus strongest control & 0.043 [0.000, 0.103]\\
Color identity / self-copy & null / exactly zero\\
\bottomrule
\end{tabular}
\end{table}


\subsubsection{Transfer in both directions}
\label{appxtransfer:task-reuse}

Directions that influence source choice within a task need not remain
equally effective on another task. To test this distinction, we compare
color-derived and material-derived directions on each task using the same
activation interchange. The direction learned on the evaluated task
provides a positive control: it shows that the interchange can change
source choice on that task.

\paragraph{Comparison.}
We use frozen Qwen2.5-7B-Instruct and estimate a separate authority direction
at each of eight layers, $10,12,\ldots,24$, from the normalized difference
between mean activations for high- and low-authority prompts in which
context and memory agree. The color directions use 200 high/low agreement
pairs from 36 factual subjects. The material directions use 48 objects
for estimation, with four high/low wording pairs per object
(192 pairs). Evaluation uses 400 color conflict pairs from those 36
subjects and 84 material objects held out from direction estimation.
Each high/low pair holds the factual subject and the asserted answer fixed.

At the final prompt position, we copy the activation component along the
tested direction from the opposite unedited authority prompt in the same
pair. Thus, even when the direction is learned on the other task, the copied
value comes from the task being evaluated. We apply the replacement after
each selected block, recomputing subsequent activations. Layers are indexed
from zero. The outcome is context-following among twelve candidate answers,
with no overlap between the color and material answer sets. We report the
same two-sided gap closure defined in the main text.

\begin{figure}[htbp]
\centering
\includegraphics[width=\linewidth]{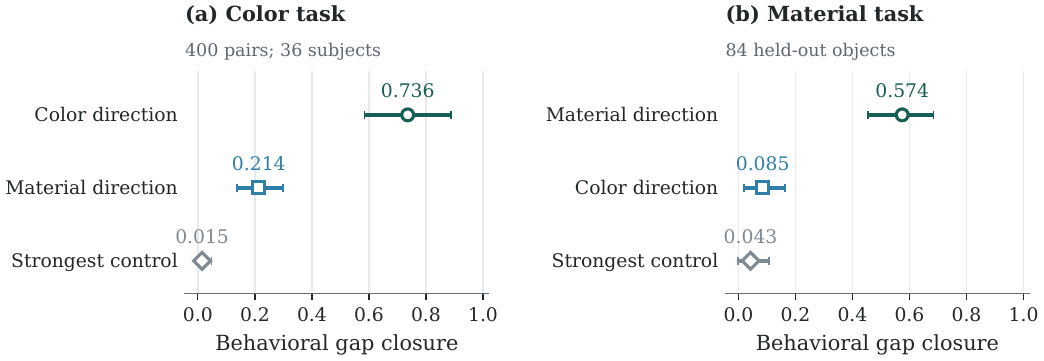}
\caption{\textbf{Directions learned on the evaluated task have larger
causal effects in both transfer directions.} Each panel compares directions
learned on that task, directions learned on the other task, and the
largest absolute closure among eight controls matched to the imported
directions' fitting data. Error bars show 95\% bootstrap intervals:
subjects with all their repeated cases for colors, and pairs within
answer-label groups for the 84 held-out material objects. Each task has
its own natural authority gap; the panels are analyzed separately.}
\label{appxtransfer:fig-task-reuse}
\end{figure}

\paragraph{Larger effects within each task.}
On the color task, color-derived directions close 0.736 of the authority
gap, compared with 0.214 for material-derived directions
(\Cref{appxtransfer:fig-task-reuse}).\footnote{The color baseline here
belongs to the task-transfer comparison; the separate cross-model analysis
reports 0.68.} On the material task,
material-derived directions close 0.574, compared with 0.085 for
color-derived directions. The paired difference in gap closure is
0.522 (95\% CI [0.389, 0.667]) for colors and
0.489 ([0.364, 0.604]) for materials. Thus the same ordering holds in
both transfer directions, with both intervals excluding zero.

The material-derived effect on colors also exceeds its strongest matched
control by 0.199 ([0.119, 0.271]). Some causal influence therefore
survives transfer, while directions learned on the evaluated task produce
substantially larger effects. These comparisons support causal use within
each task, with more limited direct reuse of learned directions across tasks.

\begin{samepage}
\paragraph{Controls and uncertainty.}
The control directions are random linear combinations of centered agreement
activations, projected orthogonally to the learned direction, and normalized
at each layer. They copy from the same prompts and use the same intervention sites.
We calculate confidence intervals from 4,000 bootstrap draws. Within
each task, the same resampled cases are used for every tested direction;
the authority gap, paired differences, and strongest control are recomputed
in each draw. For colors, we resample subjects, keeping all repeated cases
for each subject together. For materials, we resample pairs within groups
with the same correct answer. The bootstrap holds the estimated directions
fixed. Color evaluation reuses the estimation subjects, and the two tasks
use different prompts for estimation. The comparison therefore applies to
these direction estimates and cannot attribute the difference solely to
the task.

\end{samepage}

\subsection{Natural donor transfer}
\label{app:affine}

We fit directions and maps using agreement prompts for 36 color donors and 48 material calibration examples, then evaluated transfer on 84 previously evaluated material examples. For each layer, the unit directions $u_{c,\ell}$ and $u_{m,\ell}$ were the mean high minus low activation differences in the color and material banks. We projected each bank onto its direction and fit $y=a_\ell x+b_\ell$ by ordinary least squares to eight equally weighted coordinate means, one per cue and authority level, with an unconstrained slope and intercept:
\begin{equation}
 h'_{\ell}=h_{\ell,\mathrm{current}}+u_{m,\ell}
 \left(a_\ell u_{c,\ell}^{\top}h_{c,\ell,\mathrm{natural}}
       +b_\ell-u_{m,\ell}^{\top}h_{\ell,\mathrm{current}}\right).
 \label{eq:affine-operator}
\end{equation}
Here $h_{c,\ell,\mathrm{natural}}$ is the cached natural color donor activation.

Each task recipient used 8 distinct donors of the opposite authority. The source and target averages were computed by taking the mean of the four authority cues of the donor authority and applying the map to the source average. The location of the transfer was taken by referencing the natural swap locations of the materials task.

The run used the frozen Qwen model and tokenizer, BF16, eager attention, batch size 16, and no cache. Recipient projections and replacement deltas used FP32, with the delta cast to BF16 before addition. Context following records whether the context answer has the largest score among the 12 material answers. We then analyzed the saved probabilities, normalized over those same answers, including $p(\mathrm{context})$, $p(\mathrm{memory})$, and their signed difference. The reported comparisons were selected from 113 endpoints and 52 paired contrasts after the original closure result was known.

Donor slots were averaged within each recipient and authority level before using the same 10,000 paired entity bootstrap draws. The percentile 95\% intervals are pointwise and unadjusted for multiple comparisons, conditional on fixed fits and donors; they exclude uncertainty from fitting new maps or selecting new donors. Natural context following was 0.583333 for high authority and 0.023810 for low authority. Natural context answer probabilities were 0.599714 and 0.029210, compared with 0.349130 and 0.172508 after affine transfer.

\begin{table}[tb]
\centering
\caption{Gap closure and paired differences for natural donor transfer. Intervals use the same recipient draws with fits and donors fixed.}
\label{tab:affine-results}
\small
\begin{tabular}{@{}lcc@{}}
\toprule
Arm & Closure [95\% CI] & Affine minus arm [95\% CI]\\
\midrule
Affine donor & 0.684 [0.546, 0.827] & ---\\
Uncalibrated $a=1,b=0$ & 0.763 [0.607, 0.942] & -0.080 [-0.221, 0.042]\\
Mapped source average & 0.702 [0.524, 0.889] & -0.019 [-0.099, 0.063]\\
Target average & 0.766 [0.583, 0.957] & -0.082 [-0.168, 0.000]\\
Authority breaking & 0.077 [-0.088, 0.201] & 0.606 [0.468, 0.798]\\
\bottomrule
\end{tabular}
\end{table}

\begin{figure}[tb]
\centering
\includegraphics[width=\linewidth]{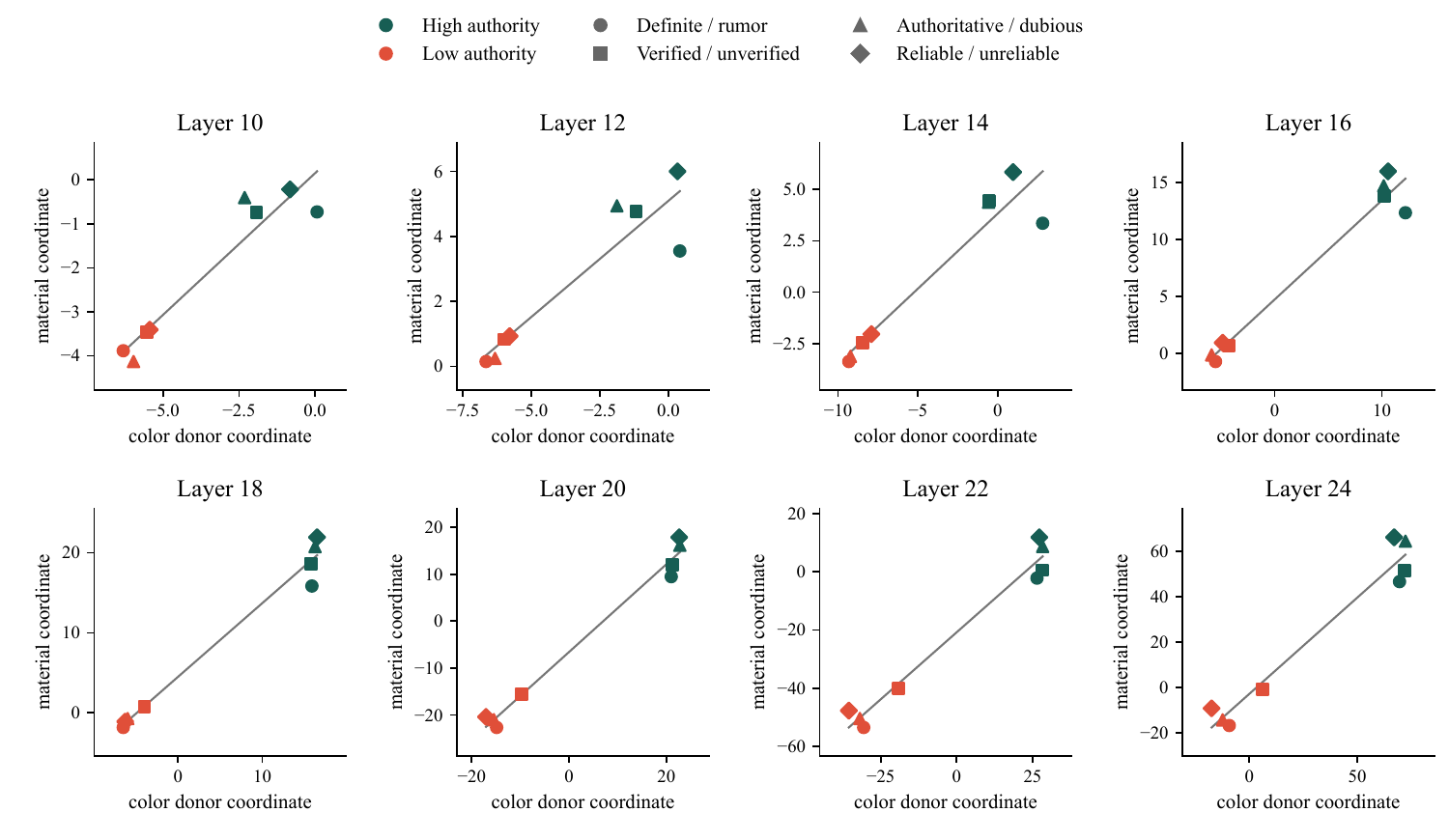}
\caption{\textbf{Calibration of natural donor coordinates.} Each panel shows eight coordinate means from the color and material calibration banks. Green marks show high authority and coral marks show low authority; marker shapes distinguish the four cue families. Gray lines show the affine fits.}
\label{fig:affine-calibration}
\end{figure}

The eight coordinate means gave $R^2$ values from 0.8587 to 0.9845 across layers (\Cref{fig:affine-calibration}). Relative to mapped source and target averages, affine transfer produced weaker suppression in high authority recipients: paired probability differences were -2.10 [-3.43, -0.83] and -2.51 [-3.86, -1.21] percentage points. Differences for low authority recipients were 0.07 [-0.52, 0.65] and 0.03 [-0.51, 0.52] points. Authority breaking also increased context answer probability in low authority recipients despite its small overall closure.

\subsection{State/country task: representation versus causal use}
\label{app:state-country}
 We formed 12 variants of facts with one or multiple token answers, each which used 32 2\texttimes{}2 fact tables. To be eligible for analysis, the variant must satisfy memory and agreement accuracy and have a sufficiently large authority gap. Only single-token state and multi-token country facts satisfied this requirement.

 The facts were split into 2,048 nonconflicting prompts, 2,048 conflicts, and 512 context-free memory prompts. For each transfer, the layer was selected by conducting four-fold table cross-validation. To address the difference in prompts between $R$ and $EA0$, a subspace $N$ is introduced in the activation computation. Specifically, $N$ is the rank-two leading uncentered right-singular subspace of paired $EA0-R$ agreement differences after projection orthogonal to the authority direction $A$. On conflicts, we evaluated natural behavior and matched single-layer interventions on $A$, $N$, and their joint subspace $A+N$. Bootstrap intervals resample the same 32 tables 10,000 times. 


\begin{table}[tb]
\centering
\caption{Reciprocal single-layer authority encoding and causal effect on state/country task.}
\label{tab:fulltransfernull}
\small
\begin{tabular}{lcc}
\toprule
Metric & countries $\leftarrow$ states & states $\leftarrow$ countries\\
\midrule
AUROC & 0.995--1.000 & 1.000\\
Encoding retained & 0.921 [0.908, 0.934] & 0.900 [0.890, 0.911]\\
Single-layer imported effect & 0.001 [-0.002, 0.004] & -0.008 [-0.011, -0.004]\\
Prompt-format interaction & -0.038 [-0.137, 0.022] & -0.049 [-0.107, 0.000]\\
\bottomrule
\end{tabular}
\end{table}


\begin{figure}[tb]
\centering
\begin{subfigure}[tb]{0.46\linewidth}
\centering
\includegraphics[width=\linewidth]{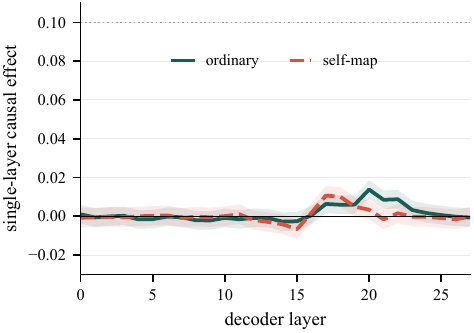}
\subcaption{Map country authority to states.}
\end{subfigure}
\hfill
\begin{subfigure}[tb]{0.46\linewidth}
\centering
\includegraphics[width=\linewidth]{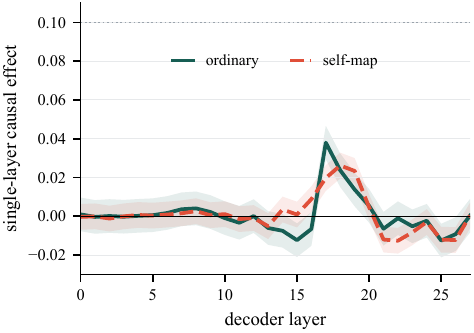}
\subcaption{Map state authority to country.}
\end{subfigure}
\caption{Causal effect of single-layer state/country swaps in Qwen2.5-7B.}
\label{fig:state-country-causal}
\end{figure}

\newpage
\end{document}